\documentclass[conference, a4paper]{IEEEtran}
\IEEEoverridecommandlockouts
\usepackage{cite}
\usepackage{amsmath,amssymb,amsfonts}
\usepackage{algorithmic}
\usepackage{graphicx}
\usepackage{textcomp}
\usepackage{xcolor}
\usepackage{stfloats} 
\usepackage{multirow}
\usepackage[OT1]{fontenc}
\usepackage{algorithmic}
\usepackage{algorithm}
\usepackage{setspace}

\def\BibTeX{{\rm B\kern-.05em{\sc i\kern-.025em b}\kern-.08em
    T\kern-.1667em\lower.7ex\hbox{E}\kern-.125emX}}

\makeatletter
\newcommand{\linebreakand}{%
  \end{@IEEEauthorhalign}
  \hfill\mbox{}\par
  \mbox{}\hfill\begin{@IEEEauthorhalign}
}
\makeatother

\renewcommand\IEEEkeywordsname{Keywords}

\begin{document}

\title{VVC Extension Scheme for Object Detection Using Contrast Reduction}

\author{\IEEEauthorblockN{Takahiro Shindo, Taiju Watanabe, Kein Yamada and Hiroshi Watanabe}
\IEEEauthorblockA{\textit{Graduate School of Fundamental Science and Engineering, Waseda University,}\\
Tokyo, Japan}

\vspace{-15pt}
}
\maketitle
\begin{abstract}
\boldmath
In recent years, video analysis using Artificial Intelligence (AI) has been widely used, due to the remarkable development of image recognition technology using deep learning. 
In 2019, the Moving Picture Experts Group (MPEG) has started standardization of Video Coding for Machines (VCM) as a video coding technology for image recognition. 
In the framework of VCM, both higher image recognition accuracy and video compression performance are required. 
In this paper, we propose an extention scheme of  video coding for object detection using Versatile Video Coding (VVC). 
Unlike video for human vision, video used for object detection does not require a large image size or high contrast.
Since downsampling of the image can reduce the amount of information to be transmitted. 
Due to the decrease in image contrast, entropy of the image becomes smaller.
Therefore, in our proposed scheme, the original image is reduced in size and contrast, then coded with VVC encoder to achieve high compression performance. 
Then, the output image from the VVC decoder is restored to its original image size using the bicubic method.
Experimental results show that the proposed video coding scheme achieves better coding performance than regular VVC in terms of object detection accuracy.
\end{abstract}

\begin{IEEEkeywords}
video coding, VVC, object detection, VCM
\end{IEEEkeywords}

\section{Introduction}
With the spread of video analysis using AI, video compression methods are essential for these purposes.
The amount of video information required for image recognition is considered to be less than that of video for human vision \cite{b1}. 
Therefore, coding schemes for image recognition that exceed the coding performance of VVC \cite{b2} are being considered.
In 2019, MPEG has positioned video coding for image recognition as VCM and started standardization.
In its standardization activities, most of the proposed coding schemes use VVC, aiming to improve coding performance by combining VVC with video pre-processing and post-processing.
In video pre-processing, some methods have been proposed to reduce the amount of information in videos. 
For example, there are pre-processing methods that convert the image size, reduce the number of frames, and use Region of Interest.
Post-processing of video includes methods such as restoring image size and converting to frames that are useful for image recognition.

However, coding scheme using image contrast reduction have not been considered, even though a lower bitrate can be achieved. 
Therefore, this paper investigates the effect of reducing the image contrast on object detection accuracy. 
For the object detection model, YOLO-v7 \cite{b4} is used, which has both high object detection accuracy and fast object detection speed.
Experimental results show that VVC extention scheme using contrast reduction reduces the bitrate required for video transmission and improves coding performance in object detection accuracy.

\section{Related Works}
VVC is the latest video coding technology standardized in July 2020.
This video coding method was created to encode video for human vision. 
However, due to its high coding performance, it has also attracted attention in VCM standardization activities.
In fact, most of the proposed video coding methods in the standardization activities partially use VVC to achieve high coding performance.

VVC reduces the amount of video information by performing intra-frame and inter-frame predictions. 
According to these prediction results, only the prediction error need to be transmitted.
After these errors are block segmented and converted into frequency components, the information mainly in the high-frequency components is removed. 
This is because the reduction of high-frequency components does not have a significant impact on image quality. 
The resulting video information from these processes is compressed to a minimal amount by entropy coding for transmission.

\section{Proposed Method}
We propose a method to lower the amount of bits required for video transmission by reducing the contrast of the image. 
Videos for object detection models need not be as vividly colored as videos for human vision. 
By lowering the contrast of video frames, the entropy of them can be suppressed. 
In other words, the bitrate required for video transmission can be reduced.
Furthermore, to decrease the amount of information in the video, the image size is halved, and the image size is restored after the video is transmitted. 
This image resizing method is described in the document published by MPEG \cite{b6}, and is utilized to create anchors of VCM.
The bicubic method is used as the image resizing method, and VVC is used as the video coding method.
The proposed video processing method is shown in Fig. 1-(a), and the algorithm used for the contrast reduction process is described below.

\begin{figure}[t]
  \centering
  \includegraphics[width=0.48\textwidth]{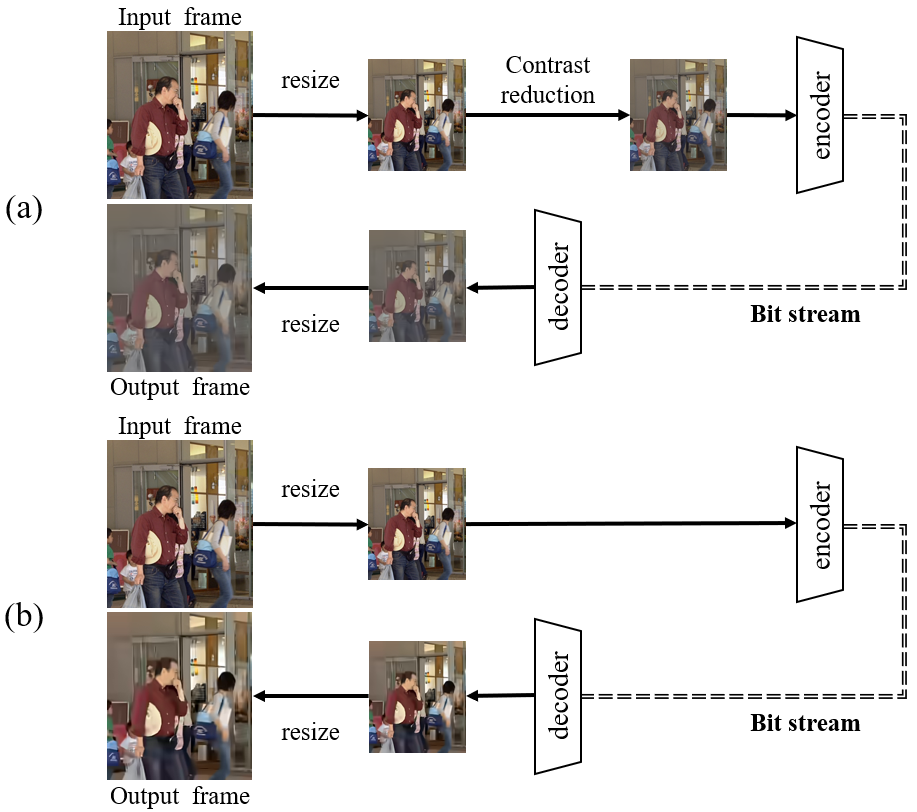}
  \caption{Processing flow of videos. (a) proposed coding scheme using contrast reduction; (b) coding scheme for comparison.}
  \label{proc}
\end{figure}

\vspace{-7pt}
\begin{equation}
  I'(x,y)=(1-\alpha)\mathalpha{*}I(x,y) + \frac{\alpha}{3\mathalpha{*}X\mathalpha{*}Y}\sum_{x=1}^X\!\sum_{y=1}^Y I(x,y),\label{eq}
\end{equation}
where $I(x,y)$ and $I'(x,y)$ indicate the input and output image.
$X$ and $Y$ are the width and height of the image.
$\alpha$ represents the ratio of the number of tones to be reduced.

\section{Experiment}

\subsection{Evaluation method}

To measure coding performance in object detection accuracy, we use the SFU-HW-v1 \cite{b5} dataset. 
This dataset consists of raw videos and the corresponding annotations for object detection, and is designated as Common Test Condition \cite{b6} in VCM standardization activities.
The dataset contains 18 raw videos, which are classified into classes A to E according to its image size and characteristics. 
In this experiment, a total of 8 sequences from classes B and C are used.

We reduce the contrast of these videos using Eq. 1. In this experiment, the value of $\alpha$ is set to 0.25.
To demonstrate the effectiveness of the proposed video coding scheme, we compare its performance with that of regular VVC. 
The processing of comparison method is shown in Fig. 1-(b). 
In order to accurately compare the encoding performance, the image size is changed as referred to in the proposed method.

VTM17.2 \cite{b7} is used as the video coding method and ``lowdelay\_P'' is used as the frame reference structure.
The quantization parameters in the proposed coding method are 14 integers from 32 to 45. 
In the coding method for comparison, the parameters are 13 integers from 35 to 47.
VVC coded video is input to YOLO-v7 to perform object detection.
The confidence threshold of the detection model is set to 0.25. 
Average Precision (AP) is used for detection accuracy, and IoU threshold is always set to 0.5 when calculating AP.

\subsection{Results}
The relationship between bitrate and mAP is shown in Fig. 2, and the relationship between bitrate and AP, which represents the detection accuracy of ``person'', is shown in Fig. 3.
In both figures, the blue curve represents the coding performance of the proposed method and the red curve represents that of the comparison method. 
By reducing the contrast of the image, it is possible to encode the image with smaller quantization parameters.
This leads to decoding images with less contrast but less block noise, which improves the coding performance in terms of object detection accuracy.

\begin{figure}[t]
\centering
\includegraphics[width=0.5\textwidth]{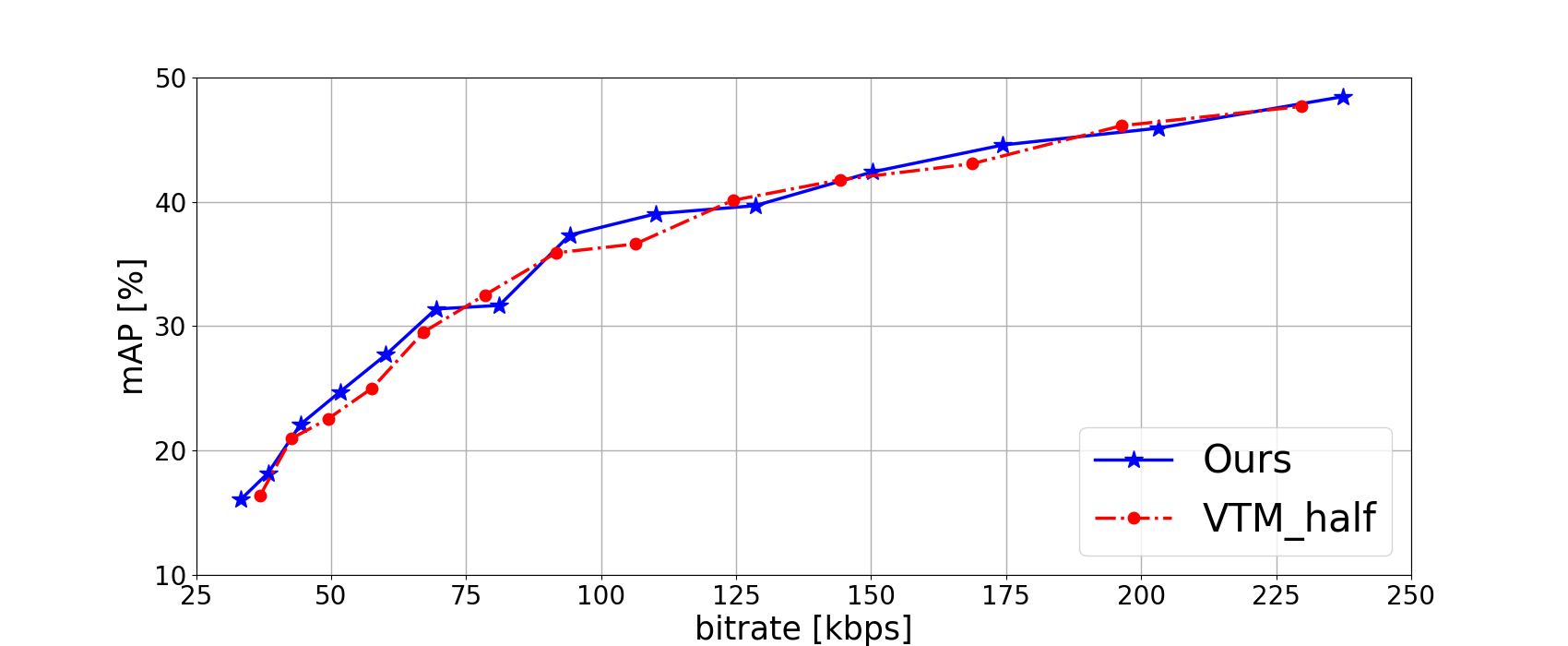}
\vspace*{-18pt}
\caption{The relationship between bitrate and mAP.}
\label{fig:mAP}
\end{figure}

\begin{figure}[t]
\centering
\includegraphics[width=0.5\textwidth]{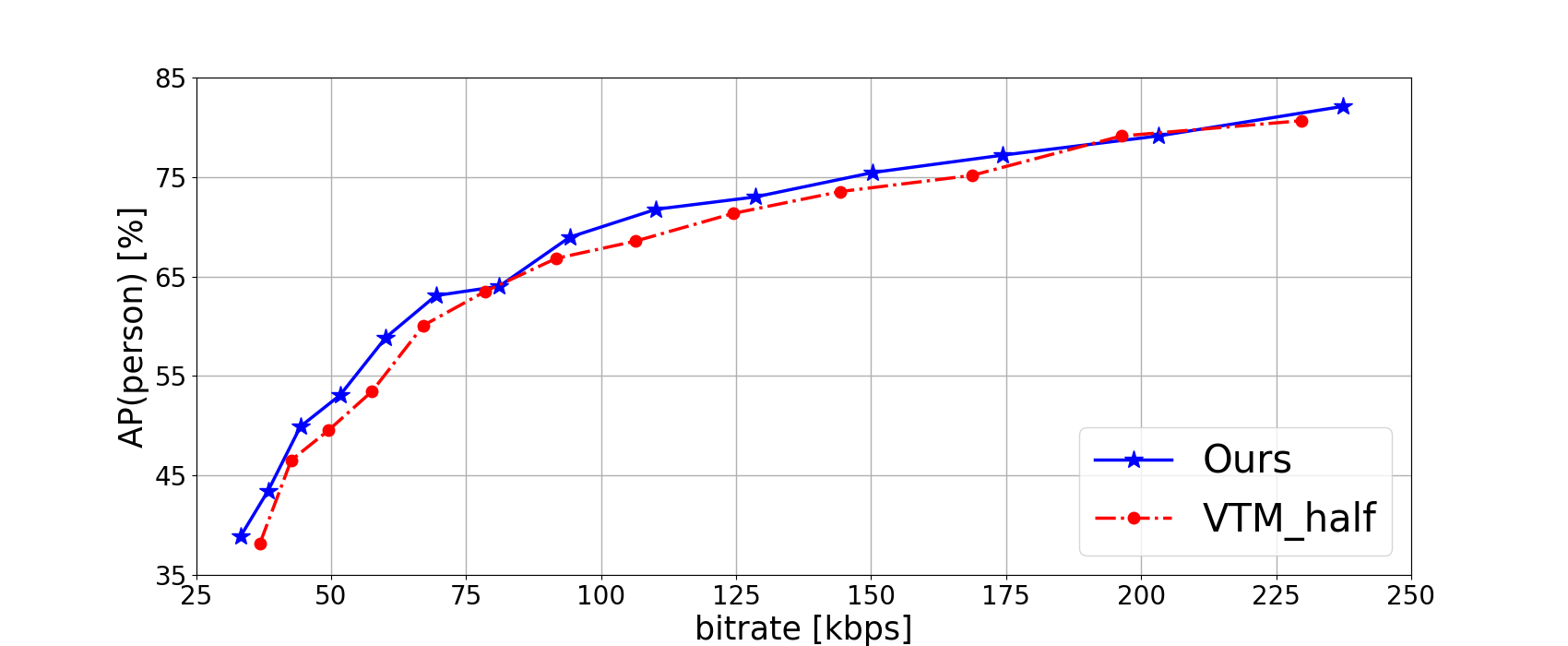}
\vspace*{-18pt}
\caption{The relationship between bitrate and AP (person).}
\label{fig:AP}
\end{figure}

\section{Conclusion}
In this paper, we proposed a coding scheme to reduce bitrate without lowering object detection accuracy by reducing image contrast before transmitting the video.
Experimental results showed that the proposed method outperforms a general VVC in terms of coding performance of object detection accuracy.
For future research, it is necessary to measure the coding performance in terms of object detection accuracy when object detection models are trained using low contrast images.

\section*{Acknowledgment}
These research results were obtained from the commissioned research (No.05101) by National Institute of Information and Communications Technology (NICT), Japan.

\end{document}